\documentclass[conference]{IEEEtran}
\IEEEoverridecommandlockouts
\usepackage{cite}
\usepackage{amsmath,amssymb,amsfonts}
\usepackage{algorithmic}
\usepackage{graphicx}
\usepackage{textcomp}
\usepackage{xcolor}
\def\BibTeX{{\rm B\kern-.05em{\sc i\kern-.025em b}\kern-.08em
    T\kern-.1667em\lower.7ex\hbox{E}\kern-.125emX}}
\begin{document}

\title{Process Mining Model to Predict Mortality in Paralytic Ileus Patients\\
{\footnotesize \textsuperscript{}}
\thanks{M. Razo, M.Pishgar, J. Theis, and H. Darabi are with University of Illinois at Chicago, Department of Mechanical and Industrial Engineering, 842 West Taylor Street, Chicago, IL 60607, United States. H. Darabi is the corresponding author. E-mail: \{mrazo3, mpishg2, jtheis3, hdarabi\} @uic.edu.
Pishgar and Razo had equal contribution to this paper. 
}
}

\author{\IEEEauthorblockN{Maryam Pishgar, Martha Razo, Julian Theis, and Houshang Darabi }

}

\maketitle

\begin{abstract}
Paralytic Ileus (PI) patients are at high risk of death when admitted to the Intensive care unit (ICU), with mortality as high as 40\%. There is minimal research concerning PI patient mortality prediction. There is a need for more accurate prediction modeling for ICU patients diagnosed with PI. This paper demonstrates performance improvements in predicting the mortality of ICU patients diagnosed with PI after 24 hours of being admitted. The proposed framework, PMPI(Process Mining Model to predict mortality of PI patients), is a modification of the work used for prediction of in-hospital mortality for ICU patients with diabetes. PMPI demonstrates similar if not better performance with an Area under the ROC Curve (AUC) score of 0.82 compared to the best results of the existing literature. PMPI uses patient medical history, the time related to the events, and demographic information for prediction. The PMPI prediction framework has the potential to help medical teams in making better decisions for treatment and care for ICU patients with PI to increase their life expectancy.

\end{abstract}


\section{Introduction}
Paralytic Ileus (PI) is the obstruction of the intestine due to paralysis of the intestinal muscles [1]. PI prevents the passage of food particles, gas, and liquids through the digestive tract leading to a backlog of food particles impairing digestive movement [2]. MediLexicon International (2018) explains the risk factors include electrolyte imbalance, advancing age, loss of weight, peripheral artery disease, and sepsis. PI is a serious condition and if prolonged and untreated will result in death. Mortality of patients with PI can be as high as $40\%$ in the ICU setting [3]. PI patients who are admitted to the ICU are especially at risk of dying because of the seriousness of their condition [4]. Early prediction of PI could be helpful for clinical decision making and effective usage of medical resources to save patients' lives. Since the results of the existing literature are limited, it is important that current research focuses on developing more accurate models for predicting mortality of ICU patients with PI to increase patients' life expectancy. 

Existing literature have developed a variety of predictive modeling to predict the mortality outcome for patients with diseases such as PI [4] and diabetes [5]. Fahad Shabbir Ahmed, et al. (2020) developed the SRML-Mortality Predictor framework with two phases. In phase 1, they performed univariate statistical analysis to filter out risk factors or variables which are not significant for predicting mortality for ICU patients with PI after 24 hours of being admitted. The authors conducted cox-regression analysis to provide the hazard ratio about the potential PI risk factors. Moreover, using the Kaplan-Meier analysis they reduced the variables to a reduced-risk factors list of 15 variables, which consists of only the statistically significant risk factors. In phase 2, Ahmed (2020) developed multiple machine learning models such as linear discriminant analysis, Gaussian naive bayes, decision tree model, k-nearest neighbor, and support vector machine (SVM) to predict mortality. SVM led to the best model performance with the AUC score of $81.38\%$. 

Besides machine learning, there is an innovative process mining framework that has shown to be promising for predicting mortality for ICU patients with diabetes [5]. In this work, demographic information of the patients, severity scores on the admission day, and timed state samples which were produced through a process mining approach called Decay Replay Mining (DREAM) and were fed into a Neural Network (NN) to predict mortality which led to an AUC of $87\%$.

In this current work, we focus on determining the mortality of ICU patients with PI after 24 hours of being admitted, using process mining modeling approach. The proposed framework is called PMPI (Process Mining Model to predict mortality of PI patients). The PMPI prediction framework demonstrates improved performance in predicting whether a PI patient dies or is discharged after 24 hours of being admitted  to the ICU.

This paper is structured as follows. Section II summarizes the preliminaries needed throughout the paper. Section III introduces the methodology. Section IV evaluates the approach by discussing the dataset and process mining model architecture and results. Section V is the discussion of the results against existing models for predicting ICU patients with PI mortality within 24 hours. Finally, we conclude the paper and discuss future work in Section VI.

\newpage
\section{Preliminaries}
The notations used in this section are adopted from [6].

\subsection{Process Mining}
Process mining is a discipline for analyzing a process using event logs. Process mining methods are classified into: Process discovery, Conformance Checking, and Process Enhancement. This paper uses process mining methods of process discovery and conformance checking. For more details on Process mining refer to [7].

\subsection{Petri Net}
A Petri Net (PN) is a mathematical model that is often used to represent a process. A PN model consists of a set of places which are graphically represented as circles. Moreover, a PN model contains transitions represented as rectangles. An example of a PN can be seen in Fig. 5. Transitions correspond to events. The occurrence of a process event is modeled by firing the transition corresponding to that event. Places can be marked or unmarked. Marked places show the current state of the process. Marking of a place is shown by putting a token (a dot) in the circle corresponding to that event. For a transition to fire, all its input places (the places with an input arrow to the transition) must have the required number of tokens. When a transition is fired, certain number of tokens are removed from all its input places. Also, each output place (any place with an input arrow from the transition) of the transition receives one or more token. For more information regarding PN the reader can refer to [7], [8] and [9].

\subsection{Event Log}
An event $a$ $\epsilon$ $A$ is an instantaneous change of a process’ state where $A$ is the finite set of all possible events [5], [6]. An event $a$ can happen more than once in a provided process. An event instance $E$ is a vector with two attributes at minimum: the name of the associated event $a$ and the corresponding occurrence timestamp, $\tau$. $E$ may contain resources. The timestamps of two events cannot be equal [5]. A trace $g$ $\epsilon$ $G$ is a finite and ordered sequence of event instances [5].

\subsection{Decay Replay Mining}
Decay Replay Mining (DREAM) [6] is a process mining based approach used to predict the next event in a process. DREAM uses the places from a discovered PN process model and extends these places with time decay functions. The time decay functions use timestamp information as a parameter when replaying an event log on a PN. Timed state samples are generated after replaying an event log on a PN that has been extended with time decay functions.

\subsection{Timed State Samples}
After replaying the event log on the PN, the timed state samples will be produced.
A timed state sample, $S(\tau)$ at time $\tau$ is the result of the concatenation of time decay function values $F(\tau)$, token counts $C(\tau)$, and of a marking $M(\tau)$:

\begin{equation}
S(\tau) =F(\tau) \oplus C(\tau) \oplus M(\tau)
\end{equation}

The timed state sample represents the state of process by containing time information [6].

\section{Methodology}
This section focuses on the PMPI framework the proposed method for predicting mortality for ICU patients with PI after 24 hours of the admission. First, the feature selection is described. Then, the conversion of Electronic Health Records (EHR) to the event logs is explained. In the last part of this section predicting the mortality of ICU patients with PI using the DREAM is introduced. A similar process mining framework was proposed by Theis for predicting in-hospital mortality of ICU patients with diabetes. The PMPI is a modified version of that framework and has been customized to handle the mortality prediction of PI patients. The PMPI prediction framework is visualized in Fig. 1.

\begin{figure}[htbp]
  \includegraphics[width=9cm, height=9cm] 
  {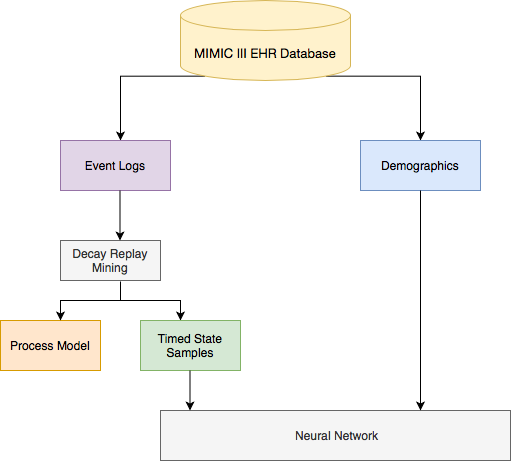}
\caption{PMPI Framework Schematic}
\label{fig:1}       
\end{figure}

\subsection{Feature Selection}
MIMIC III dataset was used for the PMPI prediction framework. To assure that every patient has a medical history available at the time of prediction, the patients who were admitted at least one time prior to the most current admission were considered. Since the goal is to predict mortality of ICU PI patients after 24 hours of being admitted, the patients who died prior to the 24 hours after being admitted to the ICU were removed. Table I shows the features that were extracted from MIMIC III. For each patient, the following features from EHR were considered: 
\begin{enumerate}
\item \textit{Admission Type:} Emergency, Elective, and Urgent.
\item \textit{Careunit in and out from ICU:} Coronary Care Unit (CCU), Cardiac Surgery Recovery Unit (CSRU), Medical Intensive Care Unit (MICU), Surgical Intensive Care Unit (SICU), and Trauma/Surgical Intensive Care Unit (TSICU).
\item \textit{Insurance Type:} Medicaid, Medicare, Private, Self Pay, and the Government.
\item \textit{Lab Items:} Creatinine, Hematocrit, Hemoglobin, Platelet, Ptt, Potassium, Aniongap, Bun, Bilirubin, Sodium, Pt, Bicarbonate, Albumin, Lactate, White Blood Count, and Glucose.
\item \textit {Age}

\item \textit{Discharge type:} Disch, Health care facility, and Dead
\end{enumerate}

\subsection{Conversion of EHR to the Event Logs}
Event logs can be understood as the sequence of events and the associated timestamp at which the events occurred. Each of the events represent preformed activities for a patient such as admission, diagnosis, lab measurements, etc., which are known as the \textit{careflows} of a patient. The patient event logs contain 51 distinct events which were created from EHR of all patients. Out of 51 distinct events, three of them belong to the admission types, 12 of them are related to the careunit activities, and represent the specific location in the ICU patients came in and out after being admitted; CCU, CRU, CSRU, MICU, SICU, and TSICU or left the aforementioned places. Moreover, 34 of the distinct events belong to the lab measurements; 17 are flagged as normal and 17 are flagged as abnormal lab measurements. Finally, two of the events are the type of patient's exit from the system, either death or discharge for each patient. Each unique event has a corresponding timestamp of when the event occurred.

\begin{table}[htbp]
\caption{Conversion of MIMIC III tables to Event Logs}
\begin{center}
\begin{tabular}{|c|c|}
\hline

\textbf{MIMIC III Tables} & \textbf{\textit{Events}} \\
\hline
{ADMISSIONS, PATIENTS} & Admission type, Demographics, \\ & patient's exit type \\
\hline

\hline
D\_ LABITEMS  & Lab events \\
\hline

\hline
ICUSTAYS, CALLOUT  &  Careunit type\\
\hline

\hline
LABEVENTS  & Abnormal flagged events \\
\hline
\end{tabular}
\label{tab1}
\end{center}
\end{table}

\begin{figure}[htbp]
  \includegraphics[width=9cm, height=5cm] {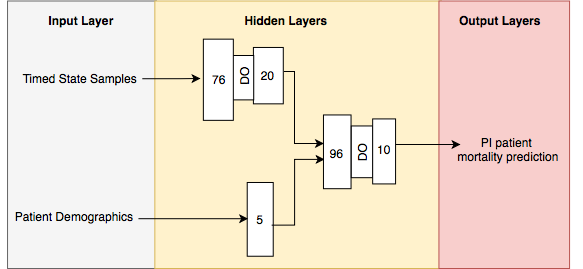}
\caption{PMPI Neural Network}
\label{fig:2}       
\end{figure}

\subsection{Prediction}
DREAM was used to predict whether a PI patient is discharged from the ICU or dies after 24 hours of being admitted. DREAM replays the event logs on a process model and produces time information which are called timed state samples.
The timed state samples, along with the demographic information of age and insurance types, are fed into the dense NN. 
The time state samples are fed into a unique branch of two hidden layers; the first hidden layer has $76$ neurons. Following the first layer, a dropout rate is defined for regularization, denoted as  $DO=0.5$. The second hidden layer contains $20$ neurons. The demographic information is fed into one single hidden layer of $5$ neurons. The aforementioned layers are concatenated into two further layers containing $96$ and $10$ neurons respectively. A dropout rate of $DO=0.5$ is defined right after the first hidden layer. An overview of the dense NN architecture for the process mining approach is visualized in Fig. 2.

\section{Evaluation}

This section discusses the experimental evaluation of the PMPI prediction mortality modeling approach for ICU patients with PI after 24 hours of being admitted. In the first subsection the dataset is described. In the following subsection the set up for modeling is described for the process model. In the third subsection, the results and comparisons to existing literature are highlighted.

\subsection{Dataset}
The data was obtained from MIMIC III (Medical Information Mart for Intensive Care). MIMIC III is a large database containing information relating to patients admitted to Beth Israel Deaconess Medical Center (BIDMC). Data includes vital signs, medications, laboratory measurements, observations and notes charted by care providers, fluid balance, procedure codes, diagnostic codes, imaging reports, hospital length of stay, survival data, and more [10]. The data set from ICU admission consisted of 46,476 total patients. A total of 1,067 PI patients were extracted using the ICD-9 code [4] from the MIMIC III database. Furthermore, the PI patients under 18 years of age at their first admission, and who died before 24 hours of being admitted to the ICU were excluded to create a final dataset of 1,017 patients.

Three data types were prepared for the PMPI prediction framework. The first data set was event logs which contained 49 unique events. These events contained EHR information about the patient including admission type, careunit type in and out from the ICU, and normal and abnormal lab items.
The second data set consisted of patient demographic data, age, and insurance. 

\subsection{Setup}
The dataset of 1,017 patients was randomly split into training and testing sets using a $67/33$ ratio producing  a train set of 681 patients and test set of 336 patients. Furthermore, the training set was randomly split using an 80/20 ratio to obtain a train and validation split. The validation sets contained 136 patients. The train and validation sets are required to discover a process model and train the NN. The train and validation set were used to select the best model and the test set was used to evaluate the model performance. 
The dataset inclusion and exclusion along with the train, test, and validation split can be visualized in Fig. 3. 
The NN has been trained for 350 epochs using a batch size of 50 with a learning rate of $5$e-$4$ and RMSprop as the optimizer. RMSprop [11] is an optimization algorithm designed for NN. The metric used was the AUC score.  AUC score is equal to the probability that a classifier will rank a randomly chosen positive instance higher than a randomly chosen negative one. AUC is a better classification estimate than other common classification performance metrics [12].
The higher the AUC, the better the model is at distinguishing between patients that are discharged and patients that die. Furthermore, the $95\%$ confidence interval (CI) for the AUC score were calculated using DeLong's method [13].

\begin{figure}[htbp]
  \includegraphics[width=12cm, height=10cm] {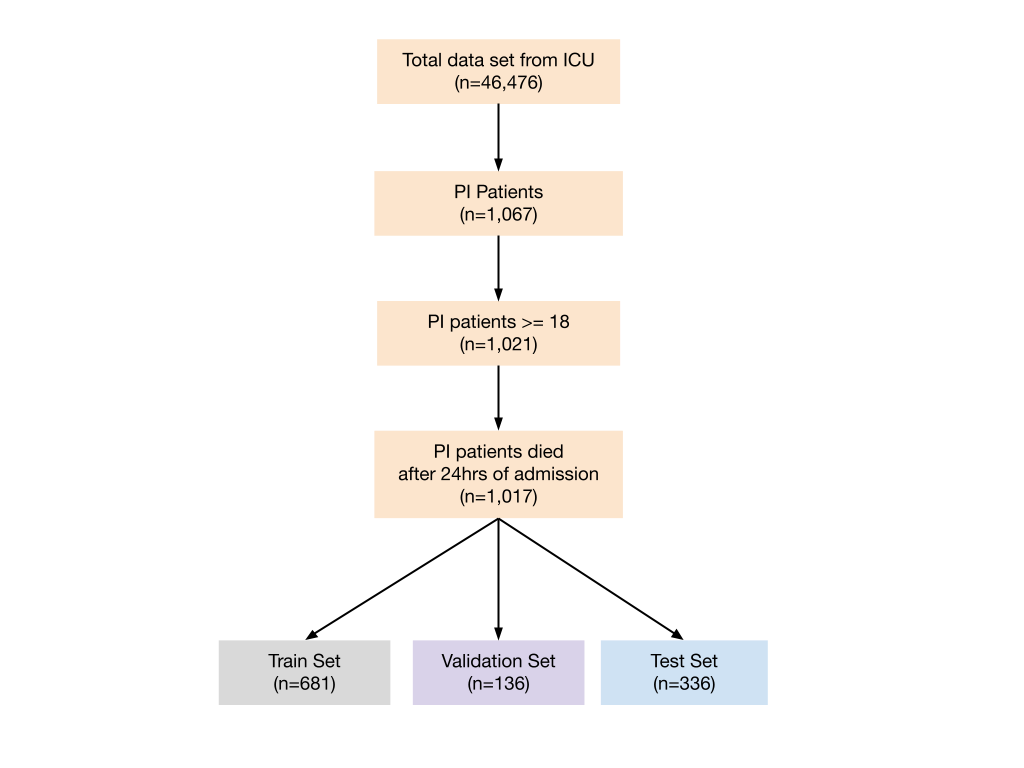}
\caption{Data utilized and data partitioning scheme}
\label{fig:3}       
\end{figure}

\subsection{Results}
The PMPI prediction approach results in the same if not better observed AUC score compared to existing literature which reports an AUC score of 0.81 [4]. The PMPI framework model resulted with an AUC of $0.820$ and $95\%$ CI of $[0.759, 869]$ using the test set. In the test set a total of 56 patients actually died, which the model was able to predict 53 of them correctly and misclassified only three of them. On the other hand, 280 patients actually got discharged, which the model was able to predict 120 of them correctly and misclassified 160 patients. 
\newline
\newline
\textit{SHAP Analysis} \newline
The impact of each feature on the model prediction is evaluated by using SHAP (SHapley Additive exPlanations) analysis [14]. The SHAP analysis for the features is shown in Fig. 4. Based on Fig. 4., demographic information has the highest impact to predict mortality for PI patients after 24 hours of admission. Lab Measurement Types follows as the second most important feature for prediction of the outcome. Lastly, Admission Types and Care Unit Types are $3rd$ and $4rd$ important features respectively. The SHAP analysis provides evidence that demographics and patients medical history information are important contributors for the model prediction.

 \begin{figure}[htbp]
  \includegraphics[width=9cm, height=5cm] {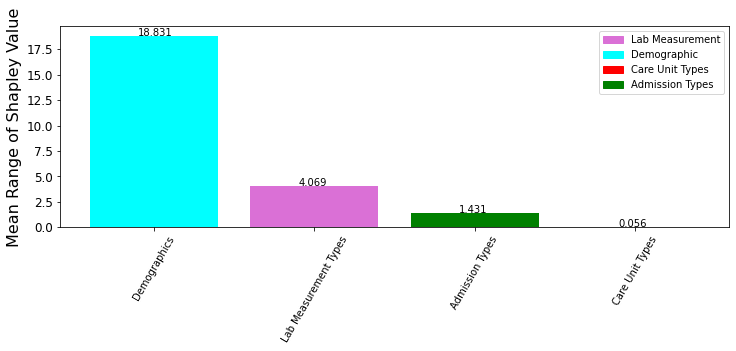}
\caption{SHAP Analysis Results}
\label{fig:4}       
\end{figure}

One of the main advantages of the process mining approach is that a process model in the form of a PN is discovered. The PN for the patients in the training dataset is visualized in Fig. 5. The PN contains a total of $22$ places and $25$ visible transitions which correspond to the events. One of the visible transitions is a merged transition. The black square represents hidden transitions which contain $24$ unique events. Hidden transitions, also referred to as invisible transitions, represent events that cannot be observed since they have little meaning to representing the process as a whole. A PN allows for visualization of patient careflows for further analysis of PI patients in an ICU setting. Based on Fig.5. the source is the yellow circle and the sink is the green circle. The source and the sink mark the admission to ICU and discharge from the system (as either dead or alive).
The unique transitions label and its associated event description are shown in Table II in the Appendix.
 \begin{figure}[htbp]
  \includegraphics[width=16cm, height=9cm] {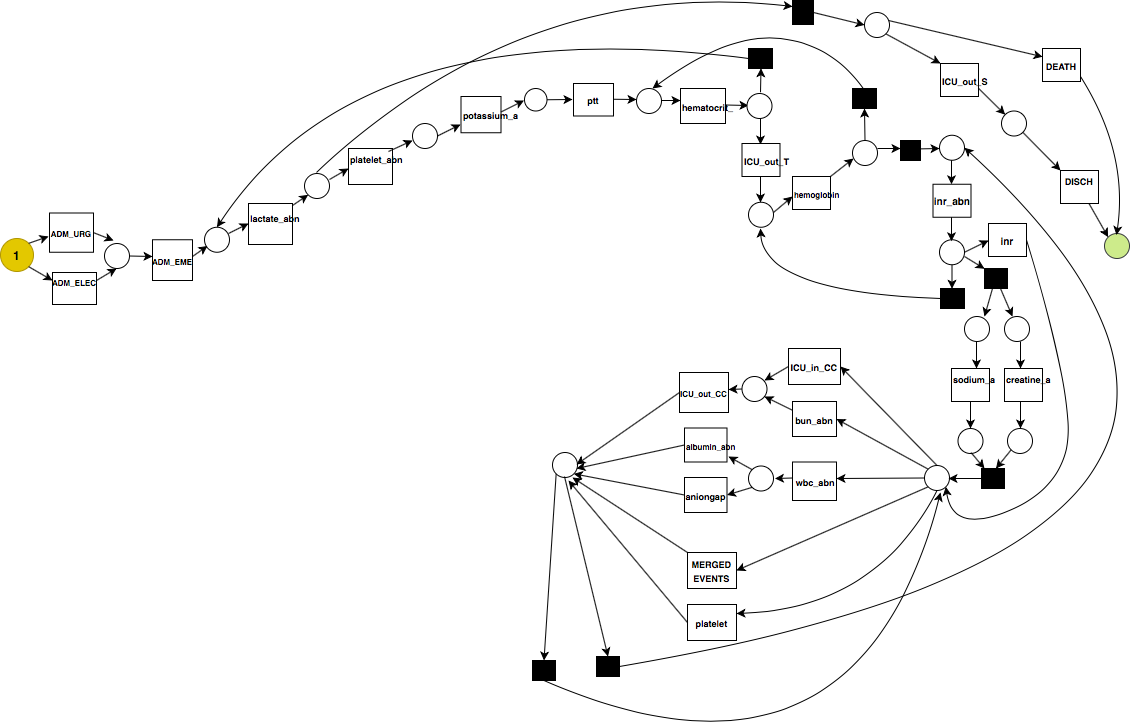}
\caption{Petri Net for ICU PI patients }
\label{fig:5}       
\end{figure}

Furthermore, process mining uses the medical history of the patients from prior hospital encounters. Also, the PMPI framework allows for incorporation of the time information related to the events, timed state samples as a feature to the NN. Existing models ignore time information; this is a useful attribute for predicting mortality of PI patients. 
A limitation of the proposed approach is that it requires patients' medical history. Unfortunately, small clinics and hospitals might not have records of patients. Also, this method will exclude new patients and patients that have not been admitted to the hospital prior to the current hospital visit. Moreover, hospitals tend not to share patient data across other networks of hospitals. Thus, the proposed approach mostly addresses large hospital networks with patients that have at least 24 hours of hospital medical data.




\section{Conclusion}
PI patients are at high risk of death when admitted to the ICU if not treated immediately. This paper demonstrates performance improvements in predicting the mortality of ICU patients diagnosed with PI after 24 hours of being admitted. The proposed framework, PMPI prediction, uses a process mining based approach. The process mining in the PMPI prediction framework shows similar if not better performance yielding an AUC score of $0.82$.
The results demonstrate the importance of using medical history and time information related to the events in order to predict and evaluate the risk of PI patients. Future work will be to explore other modeling techniques to increase the performance for predicting mortality for PI patients in the ICU after 24 hours of admission.
Furthermore, for the PMPI framework a total of 49 features were used, a future goal is to reduce the number of features for predicting mortality for ICU patients with PI. In effect, this can help medical teams use resources more effectively to increase the life expectancy of the PI patients. 

\section*{Acknowledgment}
The authors would like to thank Fahad Shabbir Ahmed and his team for providing the dataset used for their work for 

$$
$$
$$
$$
$$
$$
$$
$$
$$
$$
$$
$$
$$
$$
$$
$$
$$
$$
$$
$$
$$
$$
$$
$$
$$
$$
$$
$$
$$
$$
providing the specific subset of the MIMIC data that they used for their work in predicting mortality of PI patients. Moreover, the authors would like to thank all contributors of MIMIC-III for providing this public EHR dataset.

\newpage
\pagebreak
\section{APPENDIX}

\begin{table}[htbp]
\caption{Unique Transitions Label and Its Associated Event Discription }
\begin{center}
\begin{tabular}{|c|c|}
\hline
\textbf{Transitions Label} & \textbf{\textit{Events Description}} \\
\hline
ADM EMERGENCY  & Emergency admission \\
\hline

\hline
creatinine  &  Normal creatinine\\
\hline

\hline
glucose\_abn  & Abnormal glucose\\
\hline

\hline
hematocrit  & Normal hematocrit\\
\hline

\hline
hemoglobin  & Normal hemoglobin\\
\hline

\hline
lactate\_abn  & Abnormal lactate\\
\hline

\hline
inr\_abn  & Abnormal inr\\
\hline

\hline
platelet  & Normal platelet\\
\hline

\hline
bicarbonate\_abn  & Abnormal biocarbonate\\
\hline

\hline
ptt  & Normal ptt\\
\hline

\hline
potassium  & Normal potassium\\
\hline

\hline
aniongap  & Normal aniogap\\
\hline

\hline
pt\_abn  & Abnormal pt\\
\hline

\hline
sodium\_abn & Abnormal sodium\\
\hline

\hline
bun  & Normal bun\\
\hline

\hline
ICU\_in\_ TSICU  & Intensive  Care  Unit Trauma/Surgical\\
\hline

\hline
bilirubin  & Normal bilirubin\\
\hline

\hline
hematocrit\_abn  & Abnormal hematocrit\\
\hline

\hline
sodium  & Normal sodium\\
\hline

\hline
albumin\_abn  & Abnormal albumin\\
\hline

\hline
pt  & Normal pt\\
\hline

\hline
bun\_abn  & Abnormal bun\\
\hline

\hline
bicarbonate  & Normal bicarbonate\\
\hline

\hline
ptt\_abn  & Abnormal ptt\\
\hline

\hline
hemoglobin\_abn  & Abnormal hemoglobin\\
\hline

\hline
creatinine\_abn  & Abnormal creatinine\\
\hline

\hline
DISCH  & Discharge\\
\hline

\hline
platelet\_abn  & Abnormal platelet\\
\hline

\hline
glucose  & Normal glucose\\
\hline

\hline
potassium\_abn  & Abnormal potassium\\
\hline

\hline
ICU\_in\_MICU  & Intensive  Care  Unit Medical Intensive Care Unit\\
\hline

\hline
albumin  & Normal albumin\\
\hline

\hline
aniongap\_abn  & Abnormal aniongap\\
\hline

\hline
ICU\_in\_SICU & Intensive  Care  Unit Surgical Intensive Care Unit\\
\hline

\hline
bilirubin\_abn  & Abnormal bilirubin\\
\hline

\hline
lactate  & Normal lactate\\
\hline

\hline
ICU\_in\_CCU  & Intensive  Care  Unit Coronary Care Unit\\
\hline

\hline
ADM\_ELECTIVE  & Elective Admission\\
\hline

\hline
ICU\_in\_CSRU  & Intensive  Care  Unit Cardiac Surgery Recovery Unit\\
\hline

\hline
DEATH  & Death\\
\hline

\hline
ICU\_out\_TSICU  & Intensive  Care  Unit Trauma/Surgical Intensive Care Unit\\
\hline

\hline
wbc\_abn  & Abnormal wbc\\
\hline

\hline
inr  & Normal inr\\
\hline

\hline
ADM\_URGENT  & Urgent admission\\
\hline

\hline
ICU\_out\_CSRU  & Intensive  Care  Unit Cardiac Surgery Recovery Unit\\
\hline

\hline
ICU\_out\_MICU  & Intensive  Care  Unit Medical Intensive Care Unit\\
\hline

\hline
wbc  & Normal wbc\\
\hline

\hline
ICU\_out\_SICU  & Intensive  Care  Unit Surgical Intensive Care Unit\\
\hline

\hline
ICU\_out\_CCU  & Intensive  Care  Unit Coronary Care Unit\\
\hline
\end{tabular}
\label{tab2}
\end{center}
\end{table}

\end{document}